# Knowledge Retrieval

Vara Bhavya Sri Malli
*Department of Computer Science and Engineering*
*University Of South Florida*
Tampa, USA
varabhavyasrimalli@usf.edu

*Abstract*—Flexible task planning is still a significant challenge for robots. The inability of robots to creatively adapt their task plans to new or unforeseen challenges is largely attributable to their limited understanding of their activities and the environment. Cooking, for example, requires a person to occasionally take risks that a robot would find extremely dangerous. We may obtain manipulation sequences by employing knowledge that is drawn from numerous video sources thanks to knowledge retrieval through graph search.

## I. INTRODUCTION

Research in robotics has focused heavily on creating intelligent agents that can comprehend human intents and take action to solve issues in human-centered domains such as helping the elderly and infirm, delivering food, and cooking.

In robotic cooking, there are many instances where the robot cannot find an ingredient because of its unavailability. In order to prepare the missing items, the robot needs a task plan. Prior works introduced how FOON can be generated from video annotations and how it can be used to plan tasks. Since it contains knowledge only from the limited number of recipes. For instance, if a robot needs to prepare an omelet with a combination of ingredients that were never used together before in FOON, then it would be a problem as the concept of that type of omelet. This is especially helpful for automatically developing new FOONs and work plans for previously undiscovered recipes.

We design an iterative deepening depth search first algorithm and heuristic-based search algorithm to find a recipe from the knowledge graph. The outcomes of this paper are as follows:

1. Creating the functional units for the recipe videos
2. We develop an iterative deepening search algorithm to search the knowledge graph for the required recipes.
3. We design a heuristic-based search algorithms to find a recipe.

## II. BACKGROUND

FOON is a bipartite network that contains motion nodes and object state nodes. In general, an interactive manipulation motion of multiple objects will result in a state change from so-called input objects states to outcome objects states. Therefore, we connect the input object state nodes to the outcome object state nodes through the manipulation motion node. This arrangement would only allow the object state nodes to be connected to motion nodes and the motion nodes to be connected to object nodes, which thus forms the bipartite network.

### A. Nodes of FOON

The nodes in a bipartite FOON have two types: object sate O or motion M. Objects can also be containers of other objects (ingredients). These would cover objects such as bowls, pans or ovens which are manipulated with objects within it. For instance, in a task where the potato is chopped with a knife, both the potato and the knife are considered objects. The potato is initially in "whole" state and the knife is in "clean state". When the potato it is chopped (after the chopping motion), the states of the potato and knife are "chopped" and "dirty" respectively. In a FOON, each object node in the graph is unique in terms of its name and attributes.

### B. Edges of FOON

Due to the fact that certain nodes are the results of interactions between other nodes, a FOON is recognized as a directed graph . It's vital to notice that no two objects or motions are related to one another via edges drawn from one object node to the other or vice versa. If numerous object nodes are connected to the motion node, it implies that the objects are interacting with motion. Edges from the motion node pointing towards objects, means that those objects are the outcomes of the motion.

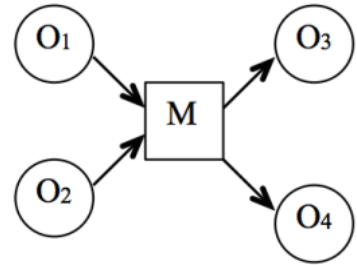

Fig. 1. A basic functional unit with two input nodes and two output nodes connected by an intermediary single motion node

### C. Functional Unit

Functional Unit is the minimum learning unit in FOON. Each unit represents a single, atomic action that is part of an activity. Each unit is a single, discrete action that makes up an activity. Input object nodes are those connected with edges pointing to the functional motion node, whereas output object nodes are those connected with edges pointing away from the functional motion node.

### D. Network Data Structure

Adjacency matrices and adjacency lists, which are common graph representations, are used to represent a FOON. The network is represented by an adjacency matrix in our work. Its ease of use both for performing and displaying a digraph network evaluation.

## III. VIDEO ANNOTATION AND FOON CREATION.

FOON can be automatically trained from observing human activities. However, given the complexity we manually input these functional units by hand, and then merge them together automatically into a single subgraph for each video. This is the reason, the creation of a FOON can be seen as a semi-automatic process

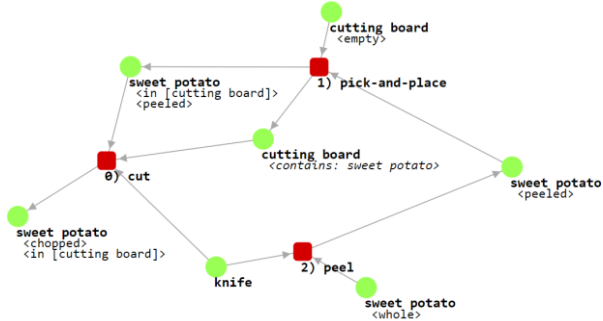

Fig. 2. A FOON subgraph based on an instructional video on making a sweet potato fry. The green solid circles are object nodes and the red solid squares are motion nodes. The object nodes are labeled with object name and their states in parentheses. The motion nodes are labeled with their manipulation motion types.

Each video's FOON subgraph is manually viewed and validated. The primary structured information required to cook the food is given in each paragraph, together with the objects (ingredients and utensils), their states, and their interactive motions. Figure 2 illustrates the FOON subgraph obtained from an online instructional video for preparing sweet potato fry.

### A. Gathering knowledge and merging

The knowledge in the FOON is derived from a collection of YouTube videos. For each video, we annotate by documenting activities along with the time they occur, changes in their states. We merge the knowledge from these subgraphs to a single bigger FOON. For merging, we conduct a union operation on all functional units while deleting duplicates. We parse these files before merging to ensure that the labels are compatible with the object and motion indices.

### B. Universal FOON

A universal FOON is defined as a merged set of two or more subgraphs from different information sources. Because a universal FOON is made up of knowledge from several sources, it can be used as a knowledge base by a robot to solve problems utilizing object-motion affordances.

### C. Knowledge Retrival

A universal FOON will provide a robot with knowledge that it can use to solve manipulation tasks given a target goal. Given specific limits, a human user may instruct a robot to create a meal. The goal of knowledge retrieval is to locate a task tree: a series of functional unit-based processes that, when completed, achieve a goal. A task tree is just a collection of functional units that are likely to be linked together and that, when executed in succession, play out the execution of steps that solve a manipulation goal. This goal can be any object node in FOON, whether it is a final product or an item in an intermediate state.

The retrieval technique for a task tree sequence is based on the concepts of fundamental graph searching algorithms; when searching, we investigate depth-wise per functional unit, but breadth-wise across items within each unit. To solve such challenges, the robot must be knowledgeable about its domain, precisely what utensils or ingredients are in its immediate surroundings, so that the system can determine whether or not a solution exists in that case. This search yields either a task tree sequence (where a target node is deemed solvable and we have a functional unit sequence that generates the goal) or no tree owing to time constraints or a non-existent solution.

As a heuristic for locating the best task tree, we use the number of units (or steps) in our search. There may be numerous units that make up an object (for example, different trees with/without the same step size), but the search method only considers the first unit that can be executed completely (or specifically, where all objects required are available as input to that unit). We can settle ties in functional units based on job complexity instead of utilizing a step-based heuristic to find a tree. A robot may be impossible to perform a given move in some cases due to constraints in its configuration space or architecture. However, we can compensate for this by performing a simpler adjustment that yields the same results. As we continue to expand FOON, we will need to make adjustments to account for other limits, such as producing meals without

## IV. METHODOLOGY

### A. Iterative Deepening Search

Iterative Deepening Search (IDS) is an iterative graph searching approach that consumes substantially less memory in each iteration while benefiting from the completeness of the Breadth-First Search (BFS) strategy (similar to Depth-First Search). IDS accomplishes the needed completeness by imposing a depth limit on DFS, which reduces the danger of becoming stuck in an infinite or very long branch. It traverses each node's branch from left to right until it reaches the appropriate depth. After that, IDS returns to the root node and explores a separate branch that is comparable to DFS.

*1) Time & space complexity:* Assume we have a tree in which each node has b children. This will be our branching factor, and d will be the tree's depth. Nodes on the lowest level, ddd, will be extended exactly once, whereas nodes on levels dld-ldl will be expanded twice. Our tree's root node will be extended d+ld+ld+l times. If we combine all of these terms, we get:

$$(d)b+(d-1)b^2+...+(3)b^{d-2}+(2)b^{d-1}+b$$

Summation of time complexity will be: $O(b^d)$
The space complexity is: $O(bd)$, In this case, we suppose b is constant and that all children are formed at each depth of the tree and saved in a stack during DFS.

*2) Time & space complexity*: It may appear that IDS has a significant overhead in the form of continuously running over the same nodes, but this is not the case. This is due to the fact that the algorithm only visits the bottom levels of a tree once or twice. Because upper-level nodes do not constitute the majority of nodes in a tree, the cost is maintained to a minimal minimum.

*3) Implementation*: we need to explore all possible paths to find the optimal solution. for making it simple, we just took the first path that we find. we kept increasing the depth until we find the solution. The task tree is considered a solution if the leaf nodes are available in the kitchen

```
1  IDS(root, goal_node, depthLmt){
2      for d = 0 to depthLmt
3          if (DepthFirstSearch(root, goal_node, d))
4              return true
5      return false
6  }
7
8  DepthFirstSearch(root, d){
9      if root == goal_node
10         return true
11     if d == 0
12         return false
13     for child in root.children
14         if (DepthFirstSearch(child, goal_node, d - 1))
15             return true
16     return false
17 }
```

### B. Greedy Best First Search

For huge search spaces, the informed search algorithm is more useful. Because informed search algorithm employs heuristics, it is also known as Heuristic search.
Heuristics function: Heuristic is a function in Informed Search that finds the most promising path. It takes the agent's current state as input and calculates how close the agent is to the goal. The heuristic method, on the other hand, may not always provide the greatest solution, but it will always discover a good solution in a fair amount of time. The heuristic function calculates how close a state is to reaching the goal. It is denoted by $h(n)$, and it computes the cost of an optimal path between two states.

**Heuristics 1**: Here we are considering the motion rates for the selecting the Input nodes as the heuristic function. The basic pseudocode follows:

**Input**: Given Goal node G and ingredients I
T ← A list of functional units in Task tree.
Q ← A queue for items to search 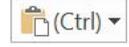
Kingd ← List of items available in kitchen.
Q.push(G)
While Q is not empty do:
    N ← Q.dequeue()
    If N not in Kingd then:
        C ← Find all functional units that create C
        max = -1
        for each candidate in C do:
            if candidate.successRate > max then:
                max = candidate.successRate
                CMax = candidate
        End for
        T.append(CMax)
        for each input in Cmax do:
            if n is not visted then:
                Q.enque(n)
                Make n visitied
        End if
        End for
    End if
End while
T.reverse()
**Output**: T

**Heuristics 2**: This algorithm is similar to the second in that the number of input nodes and their components are considered while selecting the candidate unit. A functional unit with the fewest input nodes will be picked as a candidate unit at each level.

### VII. DISCUSSION

An iterative deepening search explores the FOON by performing DFS and BFS at the chosen depth bound. The depth level will continue to rise until a solution is found. If the answer emerges at a deeper level, this approach requires more time to assemble the task tree. This will add to the temporal complexity by traversing all previously visited nodes for each depth-bound increment. Because they follow BFS, heuristics 1 and 2 easily locate the answer at higher levels, but each complexity increases if the solution occurs at deeper layers.

The task trees for all three methods could have the same or different numbers of functional units. All task trees have the same number of functional units for the target nodes ice and sweet potato.

TABLE I.  NUMBER OF FUNCTIONAL UNITS FOR TARGET NODES

| Goal Nodes | Iterative Deepening Search | Heuristics 1 | Heuristic 2 |
|---|---|---|---|

| Sweet Potato | 3 | 3 | 3 |
| Ice | 1 | 1 | 1 |
| Whipped Cream | 10 | 10 | 15 |
| Macaroni | 7 | 7 | 8 |
| Greek Salad | 31 | 32 | 28 |